\pdfoutput=1

\documentclass[11pt]{article}

\usepackage{acl}

\usepackage{times}
\usepackage{latexsym}

\usepackage[T1]{fontenc}
\usepackage[whole]{bxcjkjatype}

\usepackage[utf8]{inputenc}
\usepackage{lingmacros}
\usepackage{enumitem}
\usepackage{booktabs,multirow}
\usepackage{colortbl,array,graphicx,xcolor} 
\usepackage{tabularx}
\usepackage{microtype}
\usepackage{bbding}
\usepackage[frozencache,cachedir=.]{minted}

\definecolor{Gray}{gray}{0.935}
\newcolumntype{g}{>{\columncolor{Gray}}c}

\title{Dialogue Systems Can Generate Appropriate Responses \\ without the Use of Question Marks{\color[gray]{0.85}{?}} \\ -- An Investigation of the Effects of Question Marks on Dialogue Systems --}

\author{Tomoya Mizumoto, Takato Yamazaki, Katsumasa Yoshikawa, \\
    {\bf Masaya Ohagi, Toshiki Kawamoto, Toshinori Sato} \vspace{2mm} \\
    LINE Corporation \vspace{2mm} \\
    \texttt{tomoya.mizumoto@linecorp.com}}

\begin{document}
\maketitle
\begin{abstract}
When individuals engage in spoken discourse, various phenomena can be observed that differ from those that are apparent in text-based conversation. While written communication commonly uses a question mark to denote a query, in spoken discourse, queries are frequently indicated by a rising intonation at the end of a sentence. However, numerous speech recognition engines do not append a question mark to recognized queries, presenting a challenge when creating a spoken dialogue system. Specifically, the absence of a question mark at the end of a sentence can impede the generation of appropriate responses to queries in spoken dialogue systems. Hence, we investigate the impact of question marks on dialogue systems, with the results showing that they have a significant impact. Moreover, we analyze specific examples in an effort to determine which types of utterances have the impact on dialogue systems.

\end{abstract}

\section{Introduction} \label{sec:intro}
Given the increasing ubiquity of smartphones and smart speakers such as Siri, Alexa, and Google Assistant, voice commands have become a routine means of operating devices. In the domain of dialogue system research, some competitions have been conducted to facilitate multimodal dialogues between a user and, android robots~\cite{higashinaka-2022,minato2022-DRC} or computer graphics agents~\cite{livecompe5-eng}, that go beyond the scope of natural language processing. Numerous dialogue systems that have been used in these competitions have been constructed using textual corpora, rendering them suboptimal for spoken dialogues.

\begin{figure}[t]
\centering
\includegraphics[width=\linewidth]{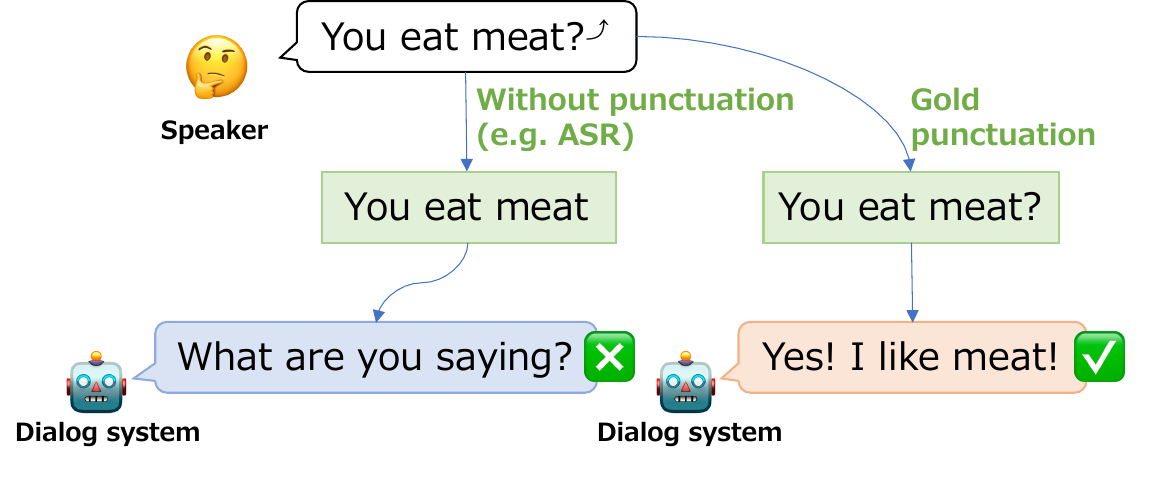}
\caption{Example of our approach to investigating how a question mark affects the quality of the response generated by a dialogue system.}
\label{fig:example}
\end{figure}

In spoken dialogue, difficulties arise that are not present in text-based dialogues.
One such issue is the lack of punctuation marks, such as question marks, in the text recognized by automatic speech recognition (ASR).
Some reports have indicated that the absence of punctuation
marks in transcribed speech can negatively impact human comprehension~\cite{Tndik2018} and performance for subsequent tasks, such as machine translation~\cite{Peitz2011}, named entity recognition~\cite{nguyen2020} and information extraction~\cite{makhoul2005}.
Efforts have been made to address this issue through research on the automatic prediction of punctuation marks, but the punctuation attachment accuracy is still around 80\%, indicating that this is a difficult task~\cite{Nozaki2022:Interspeech,zhu2022unified,li202043}.

In spoken dialogue, a rising intonation at the end of a sentence can indicate a question, but when transcribing such a sentence into text, it becomes difficult to distinguish between a declarative and an interrogative sentence.
For instance, when asking if someone is going to Prague, it is typically done through inversion, such as ``Are you going to Prague?'' in languages such as English, it is also possible to ask the question without inversion in the form of a declarative question with a rising intonation at the end, such as ``You're going to Prague?''
When transcribing these utterances using ASR, question marks and periods might be missing, which can result in existing dialogue systems designed for text-based dialogues to generating inappropriate responses.

In this study, we investigate how accurate modern dialogue systems, which are trained on written text, in responding to questions uttered without a question mark.
Specifically, we examine difference between the response generated by a dialogue system when presented with a speech input lacking a question mark, and that when presented with a correctly punctuated question (Figure~\ref{fig:example}).
Moreover, in an effort to build practical speech dialogue systems, we analyze dialogue examples where question marks are more important based on examples where response generation has failed.

\section{Research Methodology}
\subsection{Research Settings}
In this section, we discuss the research methods used to investigate the impact of question marks on dialogue systems.
To simplify the problem, we focused only on question marks, although punctuation marks such as commas, exclamation marks, and brackets are also included in text.
When dialogue is lengthy, there might be multiple sentences within a single utterance, and the number of punctuation marks such as periods and question marks increases accordingly.
However, even in cases where there are multiple sentences in a single utterance, we focus only on the final question mark in the text.

Regarding comparisons with question marks, we use two comparators: no punctuation marks and a closing period. The option without any punctuation marks is intended to simulate a scenario where the text is directly obtained from speech recognition output. The closing period is used to investigate whether there is any negative impact if automatic punctuation fails.

To generate responses, not only the target user utterance but also the past dialogue history is used as input. 
This is because it can be difficult to generate responses based solely on a brief utterance. 
In addition, in general spoken chit-chat dialogues, speakers seldom commence with a question, and thus generating a response that includes some context provides a more realistic setting. 
Therefore, in this study, we used context to generate responses, four historical utterances, including the target user utterance, used as input.

\subsection{Dialogue Data} \label{sec:data}
We used Japanese dialogue data because it is often unclear whether or not Japanese utterances are questions without question marks (e.g. [Verb]できます？'' (you/I can [Verb]?)), as mentioned in Section~\ref{sec:intro}.
Although the data were obtained from Japanese dialogues, we believe that our findings are still useful for the broader research community.

To investigate the effect of question marks in different types of dialogues, three datasets were used:  the Nagoya University Conversation Corpus~\cite{nucc} (NUCC), the dialogue history from our team's open track qualifying round in the Dialogue System Live Competition 5~\cite{livecompe5-eng} (DSLC), and the JEmpatheticDialogues (JED) dataset~\cite{sugiyama2021} .

NUCC is a collection of transcripts of casual conversations between native Japanese speakers, that have been manually transcribed by humans.
The dataset includes conversations involving multiple participants, but we only used conversations between pairs of people. 

DSLC is a collection of transcripts of chit-chat conversations between human evaluators and dialogue systems using a Large Language Model equivalent to GPT.
The evaluators and the dialogue system conducted verbal conversations, and the evaluator's speech was transcribed using the Google Speech Recognition.
Thus, we manually added periods and question marks as appropriate at the end of the speech recognition text.

JED is a collection of dialogues between two native Japanese speakers discussing an event while expressing a range of emotions, inspired by the EmpatheticDialogues~\cite{rashkin2019}.
This dataset consists of text-based dialogues, rather than spoken dialogues, and was used to compare text-based and spoken dialogues.

As noted earlier, only utterances ending with a question mark were considered in our analysis.
While the NUCC and JED datasets involve conversations between humans, for this study, we created research data by considering utterances with question marks as user utterances and the preceding utterances as the dialogue system.
Each dataset consisted of 100 dialogue examples.
Because there were less than 100 instances (only 66 instances) in the DSLC dataset where the human evaluators' utterances ended with a question mark, we augmented the sample by removing the particle ``か (ka)'' from utterances ending with interrogative particles such as ``ますか (masuka)'' or ``ですか (desuka)'' to provide more examples of utterances ending with a question mark, as shown in the following example:
\enumsentence{\label{ex:ex7}~ [Verb]できます\textbf{か}？ → [Verb]できます？ \\ (Can you [Verb]?) \hspace{1mm} → \hspace{3mm} (You can [Verb]?) \label{ex:ex1}
}

\subsection{Dialogue Systems}
We use four dialogue systems to investigate the effect of question marks.
First, we use the Transformer-based dialogue system~\cite{sugiyama2021} that won the open track of the Dialogue System Live Competition 3 held in 2020~\cite{livecompe3-eng}.
The other three dialogue systems were based on GPT~\cite{brown2020}.

The Transformer-based dialogue system pretrains a Transformer~\citep{Vaswani:17:NIPS} using Twitter reply data, and then fine-tunes it using dialogue data.
We used a model fine-tuned on JPersonaChat dialogue data~\cite{sugiyama2021}.
Hereafter, we refer to this system as ``TF''.

The GPT-based dialogue systems used OpenAI's GPT-3.5 model in addition to two in-house GPT models, one with approximately 40 billion parameters and the other with approximately 80 billion parameters. 
The in-house GPT models were trained using Japanese language data. 
The GPT-3.5 model was accessed via the API provided by OpenAI. 
These dialogue systems were developed using prompt programming (see Appendix~\ref{appendix:prompt} for an example) and are respectively referred to as the ``3.5'', the ``40B'', and the ``80B'' systems.

\subsection{Evaluation Method}
To investigate the impact of question marks on dialogue systems, we manually evaluated the responses generated by the dialogue systems.
The evaluation was conducted by four evaluators (all native Japanese speakers), with each response from a system and dataset being evaluated by three evaluators.
The evaluation was based on whether the response answered the question asked, rather than whether it was a good response.
In the interests of simplicity, we used a binary evaluation response of either good or bad.
The evaluators were given the context within which the dialogue system generated the response, and were asked to evaluate whether the response answered the question.
To prevent any biases or preconceptions based on the system or the punctuation mark, the responses were randomly presented to the evaluators in sets of 12 comprising of four systems and three punctuation marks per dialogue example, without any indication of which system or punctuation mark was used to generate the response.

To assess the agreement among the three evaluators, the Fleiss' Kappa coefficient~\cite{fleiss1973equivalence} was calculated. 
The agreement exceeded 60\% for all data, indicating a high level of agreement (see Appendix~\ref{appendix:agreement}).

\section{Results}
\begin{table}[t]
\begin{center}
\small
    \caption{Results. The
    ``?'' symbol denotes inputs that end with a question mark, the ``.'' symbols denotes inputs that end with a period, and ``X'' denotes inputs without any punctuation.}
\label{table:results}
\begin{tabular}{l|crr|crr|crr} \toprule
            & \multicolumn{3}{|c|}{\textbf{NUCC}} & \multicolumn{3}{|c|}{\textbf{DSLC}} & \multicolumn{3}{|c}{\textbf{JED}} \\ \hline
           & \multicolumn{1}{|c}{?} & \multicolumn{1}{c}{.} & \multicolumn{1}{c|}{X} & \multicolumn{1}{|c}{?} & \multicolumn{1}{c}{.} & \multicolumn{1}{c|}{X} & \multicolumn{1}{|c}{?} & \multicolumn{1}{c}{.} & \multicolumn{1}{c}{X} \\ \hline
\textbf{TF} 
& \multicolumn{1}{|r}{97}  & \multicolumn{1}{r}{61} & \multicolumn{1}{r|}{65}
& \multicolumn{1}{|r}{96}  & \multicolumn{1}{r}{67} & \multicolumn{1}{r|}{72} 
& \multicolumn{1}{|r}{98}  & \multicolumn{1}{r}{84} & \multicolumn{1}{r}{89}  \\ 

\textbf{40B} 
& \multicolumn{1}{|r}{93}  & \multicolumn{1}{r}{77}  & \multicolumn{1}{r|}{75} 
& \multicolumn{1}{|r}{97}  & \multicolumn{1}{r}{73} & \multicolumn{1}{r|}{76} 
& \multicolumn{1}{|r}{99}  & \multicolumn{1}{r}{94}  & \multicolumn{1}{r}{95} \\ 

\textbf{80B} 
& \multicolumn{1}{|r}{94}  & \multicolumn{1}{r}{83}  & \multicolumn{1}{r|}{82} 
& \multicolumn{1}{|r}{98}  & \multicolumn{1}{r}{79} & \multicolumn{1}{r|}{82} 
& \multicolumn{1}{|r}{100}  & \multicolumn{1}{r}{94}  & \multicolumn{1}{r}{98} \\ 

\textbf{3.5} 
& \multicolumn{1}{|r}{92}  & \multicolumn{1}{r}{72}  & \multicolumn{1}{r|}{76} 
& \multicolumn{1}{|r}{98}  & \multicolumn{1}{r}{86} & \multicolumn{1}{r|}{91} 
& \multicolumn{1}{|r}{99}  & \multicolumn{1}{r}{91}  & \multicolumn{1}{r}{94} \\ \bottomrule
\end{tabular}
\end{center}
\end{table}
The results of human evaluation of the dialogue systems when changing the sentence-ending punctuation are shown in Table~\ref{table:results}.
Each response was evaluated by three evaluators, and the percentage of responses that received an evaluation of ``good'' from at least two of the three evaluators is shown.

Based on these results, it is clear that responses generated with a question mark at the end achieved greater accuracy. 
Therefore, to generate accurate responses to question utterances, it is necessary to include a question mark at the end of the question.

When comparing the data types, the JED dataset was least affected by question marks, followed by the DSLC and NUCC datasets.
The JED dataset includes text-based dialogues, and often uses expressions such as ``Would it be'' that involve confirmation from the interlocutor.
Therefore, it is considered that the impact of question marks on examples from this dataset was relatively small.

Comparing the systems, the ``80B'' and the ``3.5'' systems demonstrated better performance, suggesting that larger parameters allow for the generation of more appropriate responses using dialogue context, even without question marks.

Comparing dialogue without punctuation (``X'') and with a closing 
 period (``.''), the results without punctuation were slightly better, but the difference was not significant.
 Therefore, even if a period is mistakenly added to a question utterance, the same quality of dialogue can be obtained as that using the ASR results.

\section{Discussion}

\begin{table}[t]
\begin{center}
\small
    \caption{The accuracy of the comparison based on the presence or absence of interrogative particles.}
\label{table:results2}
\begin{tabular}{l|ccc|ccc} \toprule
            & \multicolumn{3}{|c|}{\textbf{w/ interrog. particle}} & \multicolumn{3}{|c}{\textbf{w/o interrog. particle}} \\ \hline
           & \multicolumn{1}{|c}{?} & \multicolumn{1}{c}{.} & \multicolumn{1}{c|}{X} & \multicolumn{1}{|c}{?} & \multicolumn{1}{c}{.} & \multicolumn{1}{c}{X} \\ \hline
\textbf{TF} 
& \multicolumn{1}{|r}{0.97}  & \multicolumn{1}{r}{0.94} & \multicolumn{1}{r|}{0.94}
& \multicolumn{1}{|r}{1.0}  & \multicolumn{1}{r}{0.35} & \multicolumn{1}{r}{0.41} \\ 

\textbf{40B} 
& \multicolumn{1}{|r}{0.97}  & \multicolumn{1}{r}{0.94} & \multicolumn{1}{r|}{0.94}
& \multicolumn{1}{|r}{0.97}  & \multicolumn{1}{r}{0.38} & \multicolumn{1}{r}{0.44} \\ 

\textbf{80B} 
& \multicolumn{1}{|r}{1.0}  & \multicolumn{1}{r}{0.91}  & \multicolumn{1}{r|}{1.0} 
& \multicolumn{1}{|r}{1.0}  & \multicolumn{1}{r}{0.56} & \multicolumn{1}{r}{0.56} \\ 

\textbf{3.5} 
& \multicolumn{1}{|r}{1.0}  & \multicolumn{1}{r}{0.97}  & \multicolumn{1}{r|}{0.97} 
& \multicolumn{1}{|r}{0.94}  & \multicolumn{1}{r}{0.65} & \multicolumn{1}{r}{0.76} \\ \bottomrule
\end{tabular}
\end{center}
\end{table}
In this section, we analyze some actual dialogue examples.
Many of the cases where a response can be generated to a question without a question mark are those where the sentence ends with a particle indicating a question (\ref{ex:ex4}) or where there is a question-related word such as one from the ``5W1H'' (who, what, when, where, why, how) set (\ref{ex:ex5}).
Sentences for which it is difficult to generate appropriate responses without a question mark are those that even humans may perceive as declarative sentences without a question mark (\ref{ex:ex6}).
In Japanese linguistics, the subject is often omitted, making it difficult to distinguish whether a sentence is referring to ``me'' or ``you.'' 
This is one reason why it is difficult to generate responses in Japanese.
\vspace{-3mm}
\eenumsentence{
\item\label{ex:ex4}~ YouTube は見ます\textbf{か} \\ {\small (\underline{Do you} watch YouTube)} 
\vspace{-3mm}
\item\label{ex:ex5}~ \textbf{どんな}場所です \\ {\small(\underline{What} kind of place it is)}
\vspace{-3mm}
\item\label{ex:ex6}~ アメリカに行ったことある \\ {\small(I/you have been to the USA)}
}
\begin{table}[t]
    \centering
    \small
    \caption{Examples where the ``TF'' system and the ``40B'' system failed to generate appropriate responses to utterances without punctuation. The \Checkmark's signify appropriate responses are determined by human evaluators.} \label{table:exapmle2}
    \begin{tabular}{lp{5.4cm}c} \toprule
        \multicolumn{3}{c}{\textbf{Input (Context and user utterance)}} \\ \hline
        \textbf{Bot}& 失恋した友達がかわいそうでさ。 & \\
        & {\small (I'm sad for my newly single friend.)}\\ 
        \textbf{User}& えー、落ち込んでた？やっぱり。 & \\
        & {\small (Oh, they were feeling down?)}\\ 
        \textbf{Bot}&  落ち込んでたから、気分転換に行った & \\
        & {\small (They was sad, so we went out to cheer up.)} \\ 
        \textbf{User}& 友達は元気になった &\\  
        & {\small (Your friend cheers up)}\\ 
        \midrule
        \multicolumn{3}{c}{\textbf{Response of dialog system}}\\ \hline
        \textbf{TF}& それは良かったよ。 & \\ 
        & {\small(That's good to hear.)}\\ 
        \textbf{40B}& うん、それは良かった & \\ 
        & {\small(Yeah, that's good to hear.")}\\ 
        \textbf{80B}&うん、でもまだ少し暗いみたいだった & \CheckmarkBold\\ 
        & {\small (Yeah, but they still seemed a little down.)} \\ 
        \textbf{3.5}& うん、元気になってくれたみたい & \CheckmarkBold\\ 
        & {\small (Yeah, they looks like it cheered up.)}\\ 
        \bottomrule
    \end{tabular}
\end{table}

For further analysis, we investigated the impact of interrogative particles.
As described in Section \ref{sec:data}, the DSLC dataset consists of 34 instances of each condition, wherein the contexts are the same, and solely the presence or absence of a terminal interrogative particle differentiates them.
Table~\ref{table:results2} shows the performance comparison predicated on the impact of interrogative particles.
In instances where interrogative particles are present, high performance is maintained irrespective of the punctuation at the sentence's end. Conversely, in the absence of interrogative particles, the performance diminishes unless the sentence concludes with a question mark.
From this observation, it can be inferred that in cases where there is no interrogative particle, it becomes challenging to generate appropriate responses without a question mark.

Table~\ref{table:exapmle2} shows some examples of where dialogue systems could not provide appropriate responses in cases where there was no punctuation mark (see Appendix~\ref{appendix:examples} for other examples).
As mentioned earlier, it is difficult to determine whether the user's utterance is a question based solely on the utterance itself. 
In some cases, such as this example, some dialogue systems are able to generate an appropriate response by inferring a question from the context, even without a question mark.
However, the systems are unable to do so in approximately 20\% of cases. 
To solve this problem, it is not enough to use text-based information. 
Rather, it is necessary to explore methods that use non-linguistic information such as sound information or facial expressions, as  \newcite{alnuhait2023facechat} has demonstrated.

\section{Conclusion and Future Works}
In this study, we investigated the impact of a question mark at the end of a sentence on response generation in dialogue systems. 
The results showed that the presence of a question mark increased the proportion of appropriate responses by 20--30\%, indicating the significant role that question marks play in generating appropriate responses in dialogue.
Examples of failures when a question mark was absent showed that most utterances were not recognized as questions without a question mark.
Thus, the development of multimodal response generation using not only text but also sound and facial expressions is needed to address these issues.

It remains to be investigated whether periods and question marks as well as commas, in multiple sentences affect the quality of responses.
Another task for future research is to investigate whether the presence or otherwise of a question mark affects the quality of response generation in other languages.

\bibliography{custom}
\bibliographystyle{acl_natbib}

\newpage
\appendix

\label{sec:appendix}

\section{Prompt Examples} \label{appendix:prompt}
See Figure~\ref{fig:prompt}.
\begin{figure}[t]
    \centering
    \begin{minted}[breaklines,fontsize=\scriptsize,frame=single,breaksymbol=\quad,breaksymbolindentnchars=2]{text}
===  # Example of a shot
あなたはAです。AとBが雑談をします。
(You are A. A and B engage in small talk.)
以下はAとBの対話例です。
(Here are some examples of conversations between A and B.)

A:初めまして、Aといいます！さて、何のお話をしましょうか。
(A: Nice to meet you, I'm A! So, what shall we talk about?)
B:こんにちは！最近はオリンピックばかり見ていますが、Aくんは注目している競技はありますか？
(B: Hello! Recently, I've been watching a lot of the Olympics. Is there any specific event you are paying attention to?)
A:オリンピック面白いですよね。私は今回スケートボードを初めて見たのですが、とても激しくて見応えがありました。
(A: Yeah, the Olympics are really exciting! I watched skateboarding for the first time this year, and it was absolutely thrilling. )
B:スケボーですか！私も見ましたが、引き込まれましたね。他には、サッカーが今面白いですね。
(B: Oh, skateboarding! I watched it too, and it was mesmerizing. As for other sports, I've been really enjoying soccer.)
A:サッカー今面白いんですか?私は見れてないですね。今どんな感じなんですか？
(A: Is soccer exciting right now? I haven't been able to watch it. How is it going?)
B:先日日本とスペインの試合がありまして、延長戦で点数が決められて日本が負けてしまったんです。惜しかったです。
(B: The other day, there was a match between Japan and Spain, and it ended with a score in extra time, and Japan ended up losing. It was a close game.)
A:日本負けてしまったんですね。それは悔しいですね...。スペインはサッカーが強いですから、日本はよく健闘しましたね。
(Japan lost, huh? That's disappointing... Spain is known for their strong soccer skills, so Japan really put up a good fight.)
###
===  
# We repeat shots similar to that shown above.
...
===
まず、Aの発話からはじめてください。
(First, let's start with A's utterance.)
    \end{minted}
    \caption{An example of prompts used in GPT-based dialogue systems. Note that the text following the \# symbol is a comment and is not actually included in the prompt.}
    \label{fig:prompt}
\end{figure}

\section{Evaluator Agreement} \label{appendix:agreement}
\begin{table}[t]
\centering
\small
\caption{Inter-evaluator agreement among the three evaluators.}
\label{table:kappa}
\begin{tabular}{l||r|r|r} \toprule
Data & \textbf{NUCC} & \textbf{DSLC} & \textbf{JED} \\ \midrule
Agreement & 61.2 & 82.1 & 61.3     \\
\bottomrule
\end{tabular}
\end{table}
The agreement are shown in Table \ref{table:kappa}. The DSLC dataset had a Kappa coefficient of more than 80, and both the NUCC and JED dataset had a Kappa coefficient of 61, indicating high agreement.

\section{Dialogue Examples} \label{appendix:examples}
See Tables~\ref{table:exapmle1} and \ref{table:exapmle3}.
\begin{table}[t]
    \centering
    \small
    \caption{Examples where the ``TF'' system and the ``3.5'' system failed to generate appropriate responses to utterances without punctuation. } \label{table:exapmle1}
    \begin{tabular}{lp{5.4cm}c} \toprule
        \multicolumn{3}{c}{\textbf{Input (Context and user utterance)}} \\  \hline
        \textbf{Bot}&そんなの聞かなくていいです。& \\
        & {\small (You don't have to ask that.)}\\ 
        \textbf{User}&じゃあね、あの棚使ってる？ & \\
        & {\small (Are you using that shelf over there?)}\\ 
        \textbf{Bot}&あの棚、使ってますよ。& \\
        & {\small (I'm using that shelf.)} \\ 
        \textbf{User}& 組み立てられた\\  
        & {\small (I/you managed to assemble it)}\\ 
        \midrule
        \multicolumn{3}{c}{\textbf{Response of dialog system}}\\ \hline
        \textbf{TF}&組み立て大変ですよね & \\ 
        & {\small(It's difficult to assemble, isn't it)}\\ 
        \textbf{40B}&はい、組み立てられました& \CheckmarkBold\\ 
        & {\small(Yes, it has been assembled.)}\\ 
        \textbf{80B}&組み立てられたけど、大変でした& \CheckmarkBold\\ 
        & {\small (I was able to assemble, but it was difficult.)} \\ 
        \textbf{3.5}&それは良かったですね& \\ 
        & {\small (Yes, that's good to hear)}\\ 
        \bottomrule
    \end{tabular}
\end{table}
\begin{table}[t]
    \centering
    \small
    \caption{Examples where all systems except the ``3.5'' system failed to generate appropriate responses  to utterances without punctuation.} \label{table:exapmle3}
    \begin{tabular}{lp{5.4cm}c} \toprule
        \multicolumn{3}{c}{\textbf{Input (Context and user utterance)}} \\ \hline
        \textbf{Bot}& 今日はなんのお話をしましょうか？ & \\
        & {\small (What would you like to talk about today?)}\\ 
        \textbf{User}& ハイオランダって言ったことある？ & \\
        & {\small (Have you ever mentioned HiNetherlands)}\\ 
        \textbf{Bot}& すみません。わからなかったのでもう一度言ってもらえますか？ & \\
        & {\small (Sorry, I didn't understand. Could you please say it again?)} \\ 
        \textbf{User}& オランダに行ったことある &\\  
        & {\small (I/you have ever been to the Netherlands)}\\ 
        \midrule
        \multicolumn{3}{c}{\textbf{Response of dialog system}}\\ \hline
        \textbf{TF}& へー、そうなんですね & \\ 
        & {\small(Oh, I see.)}\\ 
        \textbf{40B}& オランダに行ったことあるんですね!& \\ 
        & {\small(You've been to the Netherlands! Great!)}\\ 
        \textbf{80B}& オランダに行ったことあるんですね! & \\ 
        & {\small (You've been to the Netherlands! Great!)} \\ 
        \textbf{3.5}& オランダには行ったことがあります。 & \CheckmarkBold \\ 
        & {\small (I have been to the Netherlands.)}\\ 
        \bottomrule
    \end{tabular}
\end{table}

\end{document}